\documentclass[letterpaper]{article}

\usepackage{natbib}
\usepackage{hyperref}
\usepackage{booktabs}
\usepackage[table]{xcolor}

\usepackage[ruled]{algorithm2e}
\usepackage{algorithmic}

\usepackage{lipsum}
\usepackage{soul}

\providecommand{\keywords}[1]
{
  \small	
  \textbf{\textit{Keywords---}} #1
}

\title{Problem-solving benefits of down-sampled lexicase selection}

\author{Thomas Helmuth \\
Hamilton College, Clinton, NY 13323 \\
thelmuth@hamilton.edu \\
 \\
Lee Spector \\
Amherst College, Amherst, MA 01002 \\
Hampshire College, Amherst, MA 01002 \\
University of Massachusetts, Amherst, MA 01003 \\
lspector@amherst.edu
}

\begin{document}
\maketitle

\begin{abstract}
    In genetic programming, an evolutionary method for producing computer programs that solve specified computational problems, parent selection is ordinarily based on aggregate measures of performance across an entire training set. Lexicase selection, by contrast, selects on the basis of performance on random sequences of training cases; this has been shown to enhance problem-solving power in many circumstances. Lexicase selection can also be seen as better reflecting biological evolution, by modeling sequences of challenges that organisms face over their lifetimes. Recent work has demonstrated that the advantages of lexicase selection can be amplified by down-sampling, meaning that only a random subsample of the training cases is used each generation. This can be seen as modeling the fact that individual organisms encounter only subsets of the possible environments, and that environments change over time. Here we provide the most extensive benchmarking of down-sampled lexicase selection to date, showing that its benefits hold up to increased scrutiny. The reasons that down-sampling helps, however, are not yet fully understood. Hypotheses include that down-sampling allows for more generations to be processed with the same budget of program evaluations; that the variation of training data across generations acts as a changing environment, encouraging adaptation; or that it reduces overfitting, leading to more general solutions. We systematically evaluate these hypotheses, finding evidence against all three, and instead draw the conclusion that down-sampled lexicase selection's main benefit stems from the fact that it allows the evolutionary process to examine more individuals within the same computational budget, even though each individual is examined less completely.
\end{abstract}


\section{Introduction}

Genetic programming is an evolutionary method for producing computer programs that solve specified computational problems \citep{koza:book}. When used as a supervised learning technique, genetic programming defines a problem's specifications by a set of \textit{training cases}. It then judges the ability of evolved programs to solve the problem by running each program on each training case, and measuring the distance between the program's output and the desired output. Genetic programming uses these \textit{error values} during \textit{parent selection} to determine which individuals in the population it selects to reproduce, and how many children they will produce.

The interaction between a program and the training cases is analogous to the interaction between a biological organism and the challenges presented by its environment. Organisms that are better equipped to handle these challenges have better reproductive success, and in genetic programming the programs that produce outputs closer to the desired outputs should produce more children.

Many parent selection methods have been developed for genetic programming, and they vary in the ways that they model the interactions that biological organisms have with their environments. In most, the performance of a program on all of the training cases is aggregated into a single value, referred to as a {\it fitness measure} or \textit{total error}, and the probability that a program will produce offspring is partially or entirely determined by this aggregate value.
Even multi-objective optimization methods, which select on the basis of multiple objectives, generally nonetheless aggregate performance across training cases into one objective~\citep{Deb:2002:NSGAII, Kotanchek:2006:GPTP, Kotanchek:2008:GPTP, Schmidt:2010:GPTP}. Similarly, the recent development of quality diversity algorithms~\citep{Cully2018, Cully2019} such as MAP-Elites~\citep{mouret2015illuminating, Vassiliades2018}  use aggregate fitness as part of the basis for selection.

The aggregation of performance is akin to exposing all organisms to all challenges that they could possibly face, and allowing those that perform best on average to produce more children. In biology, by contrast, each organism may face different challenges, and it will produce offspring if it survives the challenges that it happens to face before it has the opportunity to reproduce. 

The lexicase parent selection method differs from most other parent selection methods in that it avoids the aggregation of performance on different training cases into a single value \citep{Spector:2012:GECCOcompA,Helmuth:2015:ieeeTEC}.
Instead, it filters individuals by performance on training cases that are presented in different random orders for each parent selection event, with the result that different parents will be selected on the basis of good performance on different sequences of training cases. 
Additionally, children in the next generation will face different randomly shuffled cases than their parents did.
For these reasons, lexicase selection can be thought of as more faithfully modeling interactions between biological organisms and their environments.

\cite{Hernandez:2019:GECCOcomp} recently proposed two methods for subsampling the training set each generation when using lexicase selection, which were further studied by \cite{Ferguson:2019:GPTP}. \textit{Down-sampled lexicase selection} uses a different random subsample of cases each generation. \textit{Cohort lexicase selection} groups individuals into cohorts, and exposes each cohort to a different random subsample of the training cases. Both methods effectively change the environment from generation to generation by exposing individuals to different training cases. Crucially, both methods reduce the amount of computational effort required to evaluate each individual, since they run each program only on a subsample of the training cases. These computational savings can be recouped by evaluating more individuals throughout evolution. Results from \cite{Hernandez:2019:GECCOcomp} and \cite{Ferguson:2019:GPTP} indicate that both of these methods improve problem-solving performance compared to standard lexicase selection.

In this paper we concentrate on down-sampled lexicase selection, as it is simpler in concept and implementation, and both \cite{Hernandez:2019:GECCOcomp} and \cite{Ferguson:2019:GPTP} found its benefits to be comparable to cohort lexicase selection. We first conduct a more expansive benchmarking of down-sampled lexicase selection than has been conducted previously, using more benchmark problems and subsample sizes. These results confirm earlier findings that down-sampled lexicase selection produces substantial improvements over lexicase selection, and that it is robust to a range of subsample sizes.

We then turn to developing a better understanding of why down-sampled lexicase selection performs so well. One hypothesis put forward by \cite{Ferguson:2019:GPTP} is that down-sampled lexicase selection's success hinges on it enabling deeper evolutionary searches for more generations given the same computational effort. We compare this hypothesis to the hypothesis that simply evaluating more individuals in the search space is more important than deeper evolution specifically. We conduct experiments using increased maximum generations and increased population sizes (with non-increased generations), and find that they perform commensurately, indicating that deeper evolutionary lineages are not crucial to down-sampled lexicase selection's success.

We then examine the idea that by randomly down-sampling, we change the environment encountered by individuals each generation. In biology, many theorists believe that changing environments play an important role in evolutionary adaptation and speciation~\citep{levins1968evolution}. We hypothesize that changing the training cases on which down-sampled lexicase selection evaluates individuals each generation contributes to the evolvability of the system, resulting in improved performance. We test this hypothesis with an experiment that mimics down-sampled lexicase selection, except that it uses different training cases in every selection, meaning that every training case gains exposure each generation. The results of this experiment provide evidence against our hypothesis that ``changing environments'' are important for down-sampled lexicase selection.

One area where down-sampling (without lexicase selection) has proven useful is in avoiding overfitting and improving generalization, both in GP and in machine learning more generally. We explore the hypothesis that down-sampled lexicase selection's improved performance is driven by better generalization, and find that it does not hold up to the results of our experiments.

This article extends a preliminary report  that was presented at the 2020 Artificial Life conference~\citep{Helmuth:2020:ALife:downsampledlexicase}. Aside from general improvements to the clarity and completeness of the presentation in the conference paper, this article covers experiments involving more subsampling levels and more benchmark problems, with both of these extensions producing significant new results. One key area we explore is in using extremely small subsampled sets of training cases, resulting in surprisingly good performance with some notable drawbacks.

Our presentation below continues as follows: We first discuss lexicase selection and subsampling of training cases in more detail. Once we have covered these fundamental algorithms, we describe our experimental methods and present our benchmark results. We then address each of the above-described hypotheses in turn, and  conclude with our interpretation of the results and suggestions for future work.

\section{Related Work}

Unlike many evolutionary computation parent selection methods, lexicase selection does not aggregate the performance of an individual into a single fitness value~\citep{Helmuth:2015:ieeeTEC}. Instead, it considers each training case separately, never conflating the results on different cases. We give pseudocode for the lexicase selection algorithm in Algorithm~\ref{alg:lexicase}. After randomly shuffling the training cases, lexicase selection goes through them one by one, removing any individuals that do not give the best performance on each case until either a single individual or a single case remains. Lexicase selection has produced better performance than other parent selection methods in a variety of evolutionary computation systems and problem domains~\citep{Helmuth:2015:ieeeTEC, Helmuth:2015:GECCO, LaCava:EC, orzechowskiWhereAreWe2018b, Forstenlechner:2017:EuroGP, Liskowski:2015:GECCOcomp, oksanen:2017:CEC, moore:2017:ecal, moore:2018:alife, moore2019limits, moore2020specialists, aenuguLexicaseSelectionLearning2019, Metevier2019}.

\begin{algorithm}[t]
\caption{Lexicase Selection \textit{(to select a parent)}}
\label{alg:lexicase}
\begin{algorithmic}
\STATE Inputs: $candidates$, the entire population;
\STATE \phantom{Inputs:} $cases$, a list of training cases
\STATE Shuffle $cases$ into a random order
\LOOP
	\STATE Set $first$ be the first case in $cases$
	\STATE Set $best$ be the best performance of any individual in 
	    $candidates$ on \\
	    \phantom{Set} the $first$ training case
	\STATE Set $candidates$ to be the subset of
	$candidates$ that
	    have exactly $best$ \\
	    \phantom{Set} performance on $first$
	\IF {$|candidates| = 1$}
	    \STATE Return the only individual in $candidates$
	\ENDIF
	\IF {$|cases| = 1$}
	    \STATE Return a randomly selected individual from $candidates$
	\ENDIF
	\STATE Remove the first case from $cases$
\ENDLOOP
\end{algorithmic}
\end{algorithm}

\cite{Hernandez:2019:GECCOcomp} introduced down-sampled lexicase selection, a variant of lexicase selection that was developed further by \cite{Ferguson:2019:GPTP}. Down-sampled lexicase selection aims to reduce the number of program executions used to evaluate each individual by only running each program on a random subsample of the overall set of training cases, which are resampled each generation. This method reduces the per-individual computational effort, which can either be saved for decreased runtimes, or can be allocated in other ways, such as increases in population size or maximum number of generations. In order to compare with methods that do not subsample the training cases, we take the latter approach, always comparing methods equitably by limiting their total program executions per GP run.

Others have used subsampling of training data in GP, for reducing computation per individual or for improving generalization. To our knowledge, the only other work that has combined subsampling with lexicase selection besides \cite{Hernandez:2019:GECCOcomp} and \cite{Ferguson:2019:GPTP} is in evolutionary robotics, where subsampling is necessary for improving runtimes because of slow simulation speeds, though this research did not include comparisons with non-subsampled methods~\citep{moore:2017:ecal, moore:2018:alife}. Outside of lexicase selection, subsampling has been used largely to reduce the computational load of evaluating each individual, especially when considering large datasets~\citep{conf/inns/HmidaHBR16, Martinez:2017, curry:2004:CSCSI, ga94aGathercole, zhang:1999:GPIDI}. Others have proposed subsampling as a technique to reduce overfitting and improve generalization~\citep{goncalves:2013:EuroGP, Martinez:2017, Schmidt:2006:GPTP,Schmidt:2008:TEC, Schmidt:2010:gecco}. Additionally, subsampling data is common in machine learning for similar reasons (often referred to as mini-batches), as in stochastic gradient descent for improving generalization~\citep{kleinberg2018alternative}.

The work we present here, along with that of \cite{Hernandez:2019:GECCOcomp}, \cite{Ferguson:2019:GPTP} and \cite{moore:2017:ecal}, is novel in its application of subsampling when using lexicase selection, as well as applying subsampling to an already relatively small set of training data. To the latter point, many previous applications of subsampling aim to subsample a large set of example data (thousands or millions of cases) to a manageable size, say hundreds of cases. In our case, we start with a set of about 100-200 cases, and subsample to a set of 50 or less. When using a small set of $n$ training cases, lexicase selection can select parents with at most $n!$ different error vectors, since this is the number of different shufflings of cases. When $n$ is as small as 4 or 5, this limits selection to a small portion of the population, and often even less in practice. Lexicase selection typically requires 8 to 10 cases minimum to produce performance benefits, though others have successfully used it with as few as 4 cases~\citep{moore:2017:ecal}. With this in mind, it is not self-evident whether or not lexicase selection can maintain empirical benefits such as increased population diversity and problem-solving performance with such few cases. 

\section{Experimental Methods}
\label{sec:experimental-methods}

To explore the effects of down-sampled lexicase selection, we use benchmark problems from the domain of automatic program synthesis, which previous studies of down-sampled lexicase selection have used~\citep{Hernandez:2019:GECCOcomp, Ferguson:2019:GPTP}. In particular, we use problems from the ``General Program Synthesis Benchmark Suite''~\citep{Helmuth:2015:GECCO}, which require solution programs to manipulate a variety of data types and control flow structures. These problems originate from introductory computer science textbooks, allowing us to test the ability of evolution to perform the same types of programming we expect humans to perform. We use a core set of 12 problems with a range of difficulties and requirements for many of our experiments, and expand that set to 26 problems (all of the problems from the suite that have been solved by at least one program synthesis system) for one experiment. We additionally compare down-sampled lexicase selection to standard lexicase selection on the 25 problems of PSB2, the second iteration of general program synthesis benchmark problems~\citep{Helmuth:2021:GECCO:PSB2}.\footnote{More information can be found at the benchmark suite's website: \url{https://cs.hamilton.edu/~thelmuth/PSB2/PSB2.html}.}

\begin{table}[t]
    \centering
    \caption{Full training set size and program execution limit for each problem.}
    \label{table:training}
    \begin{tabular}{m{7.2cm} b{1.4cm} r}
        \toprule
        \textbf{Problems} & \textbf{Training Set Size} & \textbf{Executions} \tabularnewline
        \midrule
        Number IO & 25 & 7,500,000 \tabularnewline
        \midrule
        Sum Of Squares & 50 & 15,000,000 \tabularnewline
        \midrule
        Compare String Lengths, Digits, Double Letters, Even Squares, For Loop Index, Median, Mirror Image, Replace Space With Newline, Smallest, Small Or Large, String Lengths Backwards, Syllables & 100 & 30,000,000 \tabularnewline
        \midrule
        Last Index of Zero, Vectors Summed, X-Word Lines & 150 & 45,000,000 \tabularnewline
        \midrule
        Count Odds, Grade, Negative to Zero, Pig Latin, Scrabble Score, String Differences, Super Anagrams & 200 & 60,000,000 \tabularnewline
        \midrule
        Vector Average & 250 & 75,000,000 \tabularnewline
        \midrule
        Checksum & 300 & 90,000,000 \tabularnewline
        \bottomrule
    \end{tabular}
\end{table}

As in~\cite{Helmuth:2015:GECCO}, we define each problem's specifications as a set of input/output examples, so that GP has no knowledge of the underlying problems besides these examples.\footnote{Datasets for these problems can be found at \url{https://git.io/fjPeh}.} For each problem we use a small set of \textit{training cases} to evaluate each individual: between 25 and 300 cases per run (see Table~\ref{table:training}) and 200 cases for every problem in PSB2. We use a larger set of unseen \textit{test cases}, which are used to determine whether an evolved program that passes all of the training cases generalizes to unseen data. Before testing a potential solution for generalization, we use an automatic simplification procedure that has been shown to improve generalization~\citep{Helmuth:2017:GECCO}; finding a simplified program that passes all of the unseen test cases is considered a successful GP run. We test the significance of differences in numbers of successes between sets of runs using a chi-square test with a 0.05 significance level, using Holm's correction for multiple comparisons whenever there are more than two methods run on a single problem in one experiment.

When a run using down-sampled lexicase selection finds a program that passes all of the subsampled training cases, we do not immediately terminate the run. Instead, we run the program on the full training set (using it as a validation set), and terminate the run if the program passes all of those cases. If it does not, we continue to the next generation, as the individual (or its children) may not pass some of the cases in the newly subsampled set of cases.  As we will detail in Section~\ref{sec:lower-bound-subsampling}, with an extremely low subsampling level that leaves the subsampled training set with 1 or 2 cases, it is easier for GP to generate individuals that perfectly pass those cases without passing the full training set; with enough of these individuals, the process of verifying that they pass the full training set may dominate the running time of evolution. Note that if only a single individual passes all cases in the subsampled training set but evolution continues, it will receive every single parent selection in that generation. These \textit{hyperselection events}~\citep{Helmuth:2016:GECCO} may have strong effects on population diversity, a potential avenue for future study.

We evolve programs with the PushGP genetic programming system, which uses programs represented in the Push programming language~\citep{1068292, spector:2002:GPEM}. Push was designed with genetic programming in mind, in particular to enable \textit{autoconstruction}, in which evolving programs not only need to try to solve a problem, but are also run to produce their children~\citep{spector:2002:GPEM,Spector:2016:GECCOcompA}. 
Push programs utilize a handful of typed stacks, from which instructions pop their arguments and to which instructions push their results. 
Push programs can be any hierarchically-nested list of instructions and literals, the latter of which the interpreter pushes onto the relevant stack. We use the Clojush, the Clojure implementation of PushGP, for our experiments.\footnote{\url{https://github.com/lspector/Clojush}}

\begin{table}[t]
    \centering
    \caption{PushGP system parameters.}
    \label{table:parameters}
    \begin{tabular}{l r}
        \toprule
        \textbf{Parameter} & \textbf{Value} \tabularnewline
        \midrule
        population size & 1000 \tabularnewline
        max generations for runs using full training set & 300 \tabularnewline
        genetic operator & UMAD \tabularnewline
        UMAD addition rate & 0.09 \tabularnewline
        \bottomrule
    \end{tabular}
\end{table}

We present the PushGP system parameters used in our experiments in Table~\ref{table:parameters}. Our only genetic operator, uniform mutation with additions and deletions (UMAD), adds random genes before each gene in a parent's genome at the \textit{UMAD addition rate}, and then deletes random genes at a rate to remain size-neutral on average. We use UMAD to produce 100\% of the children, instead of also using a crossover operator, since thus far it has produced the best results of any operator tested on these problems~\citep{Helmuth:2018:GECCO}.

Each problem in the benchmark suite prescribes a number of training cases to use~\citep{Helmuth:2015:GECCO}. In our default configuration, we run every individual on every training case, meaning the total number of program executions allowed in one GP run is the number of training cases multiplied by the population size and generations. Since our down-sampled lexicase selection experiments use fewer cases to evaluate each individual, we limit our GP runs by a program execution limit, as given in Table~\ref{table:training}, to ensure that each method receives equal training time.

\section{Benchmarking Down-sampled Lexicase Selection}

In the work introducing down-sampled lexicase selection, experiments benchmarked down-sampled lexicase selection with subsampling levels of 0.05, 0.1, 0.25, and 0.5 on five program synthesis problems~\citep{Hernandez:2019:GECCOcomp, Ferguson:2019:GPTP}. We expand on those benchmarks by testing 3 additional subsampling levels, 0.01, 0.02, and 0.175, with the first two explicitly trying to gauge how low the subsampling rate can get before having deleterious effects. Our experiments increase the number of benchmark problems to 12, and additionally test the subsampling level of 0.25 on 39 other program synthesis benchmark problems to broaden our assessment. As described above, our experiments use PushGP, showing that the benefits of down-sampling generalize beyond the linear GP system used in the initial experiments~\citep{Hernandez:2019:GECCOcomp, Ferguson:2019:GPTP}.

\begin{table}[t]
    \centering
    \caption{Number of successes out of 100 GP runs of down-sampled lexicase selection with proportional increases in maximum generations per run across seven different subsampling levels, as well as 1.0, which is equivalent to standard lexicase selection. The mean rank calculates the average rank of each method among all methods across the problems, excluding Mirror Image and Smallest, easy problems where results differ only in random changes in solution generalization. For 6 sets of runs at subsampling levels 0.01 and 0.02, we were not able to finish all 100 runs, as described in Section~\ref{sec:lower-bound-subsampling}; the number of finished runs is given after the /.}
    \label{table:down-sample}
    \rowcolors{2}{gray!15}{white}
	\begin{tabular} {l rrrrrrrr}
    	\toprule
    	 & \multicolumn{8}{c}{Subsampling Level} \\
        Problem          & 0.01 & 0.02 & 0.05 & 0.1 & 0.175 & 0.25 & 0.5 & 1.0 \\
        \midrule
        CSL              & 0/4    & 48/97   & 38   & 25  & 60    & 51   & 40  & 32  \\
        Double Letters   & 5/42    & 85   & 87   & 72  & 55    & 50   & 29  & 19  \\
        LIOZ             & 94   & 90   & 72   & 68  & 61    & 65   & 63  & 62  \\
        Mirror Image     & 100  & 100  & 100  & 99  & 99    & 99   & 100 & 100 \\
        Negative To Zero & 96   & 84   & 84   & 86  & 86    & 82   & 78  & 80  \\
        RSWN             & 100  & 100  & 99   & 96  & 97    & 100  & 93  & 87  \\
        Scrabble Score   & 1/98    & 7    & 18   & 19  & 24    & 31   & 28  & 13  \\
        Smallest         & 100  & 100  & 100  & 99  & 100   & 98   & 100 & 100 \\
        SLB              & 100  & 100  & 99   & 96  & 96    & 95   & 94  & 94  \\
        Syllables        & 11/60   & 47   & 48   & 61  & 68    & 64   & 54  & 38  \\
        Vector Average   & 100  & 100  & 100  & 98  & 99    & 97   & 95  & 88  \\
        X-Word Lines     & 25/60   & 96   & 98   & 95  & 94    & 91   & 86  & 61  \\
        \midrule
        Mean           & 61.0 & 79.8 & 78.6 & 76.2 & 78.3 & 76.9 & 71.7 & 64.5 \\
        Mean Rank      & 4.8  & 3.2  & 3.4  & 4.2  & 3.7  & 4.0  & 5.8  & 7.1 \\
        \bottomrule
    \end{tabular}
\end{table}

\subsection{Subsampling Levels}

Table~\ref{table:down-sample} presents the success rates for down-sampled lexicase selection using seven different subsampling levels across twelve representative benchmark problems, along with the mean number of successes. The last column of 1.0 performs no down-sampling, and therefore represents standard lexicase selection. For these runs, we proportionally increase the maximum number of generations that evolution can run to keep a constant number of program executions; for example, while standard lexicase selection runs for at most 300 generations, the runs with a subsampling level of 0.02 run for at most $\frac{1}{0.02} = 50$ times as many, at 15000 generations.
For each problem, we calculate the rank of each subsampling level, and average those to calculate the mean rank, where lower values are better. Six sets of runs (five at subsampling level of 0.01 and one at level 0.02) were not able to complete in a reasonable amount of time, as discussed in Section~\ref{sec:lower-bound-subsampling}.

The subsampling level of 0.02 performed the best on average, propelled by its significantly better results on the difficult Double Letters and Last Index of Zero problems.
However, every subsampling level performed well, and all considerably better than standard lexicase (i.e. subsampling level of 1.0). The level of 0.5 performed worst of the subsampling levels, likely because it only runs for twice as many generations as standard lexicase selection, where the other subsampling levels run longer.

It is surprising that the subsampling level of 0.02 performed best, as it only uses 2 training cases per generation for seven of the problems, significantly limiting the information contained in the errors on which lexicase selection bases selection. In fact, with only 2 training cases, lexicase selection can only select individuals with 2 different error vectors corresponding to the 2 possible orderings of the cases!
Even so, this extreme constraint on selection introduced by down-sampling seems to be largely outweighed by increasing the maximum number of generations manyfold.


Even though a subsampling level of 0.02 performed best,  subsampling levels 0.02, 0.05, 0.1, 0.175, and 0.25 performed nearly identically, showing that down-sampled lexicase selection is robust to wide a variety of subsampling levels across an order of magnitude.

\begin{table}[tbp]
    \centering
    \caption{Number of successful runs comparing lexicase selection to down-sampled lexicase selection with a subsampling level of 0.25 on 26 benchmark problems. Underlined values indicate significant improvement of down-sampled lexicase over lexicase using a chi-squared test. Lexicase was never significantly better than down-sampled lexicase. ``problems solved'' counts the number of problems each method solved at least once.}
    \label{table:lex-down-sample}
    \rowcolors{2}{gray!15}{white}
	\begin{tabular} {l rr}
    	\toprule
        Problem                    & Down-sampled & Lexicase \\
        \midrule
        Checksum                   & \underline{18} & 1           \\
        CSL                 & \underline{51} & 32                  \\
        Count Odds                & 11 & 8                   \\
        Digits                      & 28 & 19                \\
        Double Letters                 & \underline{50} & 19              \\
        Even Squares                    & 2   & 0            \\
        For Loop Index                & 5     & 2            \\
        Grade                           & 2    & 0           \\
        Last Index of Zero            & 65   & 62            \\
        Median                          & 69   & 55          \\
        Mirror Image                  & 99   & 100           \\
        Negative To Zero              & 82    & 80           \\
        Number IO                      & 99   & 98           \\
        Pig Latin                    & 0   & 0               \\
        RSWN                      & \underline{100}       & 87         \\
        Scrabble Score               & \underline{31}    & 13            \\
        Small Or Large                  & \underline{22}   & 7           \\
        Smallest                      & 98   & 100           \\
        String Differences             & 1   & 0            \\
        SLB                          & 95     & 94            \\
        Sum of Squares                & 25    & 21           \\
        Super Anagrams             & 4        & 4            \\
        Syllables                    & \underline{64}     & 38           \\
        Vector Average              & \underline{97}      & 88           \\
        Vectors Summed               & 21    & 11           \\
        X-Word Lines                  & \underline{91}    & 61           \\
        \midrule
        problems solved    &  25 &  22 \\
        \bottomrule
    \end{tabular}
\end{table}

\subsection{Lower bounds of subsampling level}
\label{sec:lower-bound-subsampling}

Since down-sampled lexicase selection performs well at quite low levels of subsampling, are there any drawbacks? Is there a lower bound to the benefits of subsampling?

First, we will examine the results on our lowest subsampling level, 0.01. We see that it performed excellently on 7 out of 12 problems, including four where it operated on a single training case (Mirror Image, Replace Space with Newline, String Lengths Backwards, and Smallest) and three others with only 2 or 3 cases. These results include producing the absolute best results on two problems, Last Index of Zero and Negative to Zero. However, it gave polarized performance, producing the worst results on the remaining 5 problems. For two of these problems (Scrabble Score and Syllables), there is a clear trend toward worse performance with the lowest subsampling levels, but for the other three, down-sampled lexicase selection performs well even at the 0.02 subsampling level. We interpret these findings to suggest that, at least for some problems, 1 to 3 cases is not sufficient information to drive evolution toward solutions, likely resulting in either catastrophic lack of diversity, thrashing of the population between trying to solve different cases, or other detriments.

Beyond the problem solving performance considerations, using extremely low subsampling levels results in other unwanted behaviors of the GP system. Typically in GP, we consider the program executions to be the time limiting factor of running GP, and therefore tune our experiments to use the same number of program executions regardless of down-sampling. However, as we proportionally increase the number of maximum generations to make up for fewer program executions per generation, the remaining components of the GP system (such as genetic operators and data logging) take up a larger proportion of the running time in practice. Additionally, if we run evolution for many generations (for example 100 times as many with subsampling level of 0.01), we will require that many times more hard drive space to log data from runs. Similar issues exist with reserving sufficient RAM when increasing the population size instead of the maximum generations.

In Table~\ref{table:down-sample}, six of our sets of runs at low subsampling levels were not able to finish all 100 runs in a reasonable amount of time, and were cutoff before finishing. Some of the extreme length of these runs is likely attributable to the effects discussed in the previous paragraph. However, a subtler and potentially more harmful effect is at play as well. As described in Section~\ref{sec:experimental-methods}, when GP finds a program that passes all of the subsampled training cases, we must test it on the remaining training cases before calling it a potential solution and halting evolution; if it does not pass all training cases, evolution continues. With extremely small subsampled sets, it becomes easier for evolution to find (many) individuals that pass all of the subsampled data, requiring us to fully evaluate those individuals, which many times do not pass the full training set. This problem is compounded for problems that have Boolean outputs (such as Compare String Lengths), since even if the entire population chooses between \texttt{True} and \texttt{False} randomly, if there is only 1 case in the subsampled set, half of the population will answer that case correctly and need to be evaluated on every training case every generation, negating the benefits of quick evaluation per generation. This certainly impacted the low number of finished runs of the Compare String Lengths problem at the 0.01 subsampling level, and likely contributed to unfinished runs on other problems at that level.

With these drawbacks in mind, we see subsampling levels between 0.05 and 0.25 producing good compromises between problem solving performance and real running times. In the following section, we benchmark down-sampled lexicase selection using a subsampling level of 0.25, though we expect the results would look similar at a variety of subsampling levels.

\subsection{Expanding benchmarking of down-sampled lexicase selection to more problems}

\begin{table}[tbp]
    \centering
    \caption{Number of successful runs comparing lexicase selection to down-sampled lexicase selection with a subsampling level of 0.25 on the 25 new benchmark problems of PSB2. Underlined values indicate significant improvement of down-sampled lexicase over lexicase using a chi-squared test. Lexicase was never significantly better than down-sampled lexicase. ``problems solved'' counts the number of problems each method solved at least once.}
    \label{table:psb2}
    \rowcolors{2}{gray!15}{white}
	\begin{tabular} {l rr}
    	\toprule
        Problem            & Down-sampled        & Lexicase  \\
        \midrule
        Basement                 & 2    & 1                     \\
        Bouncing Balls           & 3    & 0                     \\
        Bowling                   & 0     & 0                   \\
        Camel Case                & 4    & 1                    \\
        Coin Sums                & \underline{39}    & 2                    \\
        Cut Vector               & 0    & 0                     \\
        Dice Game               & 1    & 0                      \\
        Find Pair               & \underline{20}   & 4                      \\
        Fizz Buzz                & \underline{74}  & 25                     \\
        Fuel Cost              & \underline{67}    & 50                     \\
        GCD                      & \underline{20}     & 8                   \\
        Indices of Substring      & 4   & 0                     \\
        Leaders                  & 0     & 0                    \\
        Luhn                 & 0        & 0                     \\
        Mastermind           & 0        & 0                     \\
        Middle Character         & \underline{79}   & 57                    \\
        Paired Digits            & 17    & 8                    \\
        Shopping List        & 0        & 0                     \\
        Snow Day                 & 7    & 4                     \\
        Solve Boolean        & 5        & 5                     \\
        Spin Words           & 0        & 0                     \\
        Square Digits           & 2     & 0                     \\
        Substitution Cipher     & \underline{86}   & 61                     \\
        Twitter              & \underline{52}      & 31                     \\
        Vector Distance      & 0        & 0                     \\
        \midrule
        problems solved   &  17 &  13  \\
        \bottomrule
    \end{tabular}
\end{table}

After extensively testing a variety of subsampling levels on 12 benchmark problems, we want to exhibit its performance on a larger set of benchmark problems. We only had the computational resources to test one subsampling level on this larger set of problems, and chose 0.25. While the subsampling level of 0.25 did not produce the best results in Table~\ref{table:down-sample}, it performed almost as well as any level, and was less computationally demanding than much lower subsampling levels for the reasons discussed in Section~\ref{sec:lower-bound-subsampling}.

Table~\ref{table:lex-down-sample} compares standard lexicase selection (i.e. the column 1.0 in Table~\ref{table:down-sample}) to down-sampled lexicase selection with a subsampling level of 0.25 on 26 benchmark problems from \cite{Helmuth:2015:GECCO}, including the 12 from Table~\ref{table:down-sample}.
Down-sampled lexicase selection produced significantly more successful runs than lexicase selection on 9 out of the 26 problems. It additionally found solutions to 3 of the problems that lexicase selection never solved, and had fewer successes on only two of the problems, neither of which were significantly different.

Table~\ref{table:psb2} continues the comparison from Table~\ref{table:lex-down-sample} on 25 new problems from PSB2~\citep{Helmuth:2021:GECCO:PSB2}. These problems were designed to be a step more difficult than those from \cite{Helmuth:2015:GECCO}, and show lower success rates for both standard lexicase selection and down-sampled lexicase selection. However, down-sampled lexicase selection continues to clearly outperform standard lexicase selection, solving 4 problems that standard lexicase never solved, and performing significantly better on 8 of the problems. In fact, down-sampled lexicase never produced fewer solutions than standard lexicase on any of the 25 problems.
This expanded benchmarking confirms previous findings that down-sampled lexicase selection creates great improvements in performance compared to lexicase selection.

\begin{table}[t]
    \centering
    \caption{Number of successful runs comparing down-sampled lexicase at a 0.1 subsampling level (DS 0.1) to using lexicase selection with a static set of 10 random training cases, which do not change during evolution. Underlined successes are significantly better using a chi-squared test.}
    \label{table:static}
    \rowcolors{2}{gray!15}{white}
	\begin{tabular} {l rr}
    	\toprule
        Problem                    & DS 0.1 & Static \\
        \midrule
        Compare String Lengths     & \underline{25}  & 0      \\
        Double Letters             & \underline{72}  & 4      \\
        Last Index of Zero         & \underline{68}  & 7      \\
        Mirror Image               & \underline{99}  & 13     \\
        Negative To Zero           & \underline{86}  & 31     \\
        Replace Space with Newline & \underline{96}  & 57     \\
        Scrabble Score             & 19  & 13     \\
        Smallest                   & \underline{99}  & 40     \\
        String Lengths Backwards   & \underline{96}  & 35     \\
        Syllables                  & \underline{61}  & 9      \\
        Vector Average             & \underline{98}  & 71     \\
        X-Word Lines               & \underline{95}  & 35     \\
        \bottomrule
    \end{tabular}
\end{table}

\subsection{Comparison with static subsample of cases}

One question raised by~\cite{Ferguson:2019:GPTP} is whether down-sampled lexicase selection's method of randomly replacing the subsampled training cases each generation is beneficial, or if a static subsample of training cases would be just as good. To examine this question, we performed a set of runs that uses lexicase selection with a static, randomly subsampled set of 10 training cases that do not change during evolution; this uses an increased number of maximum generations like with down-sampled lexicase selection. Since each problem uses a different number of training cases (100 or 200 for most benchmark problems), this is not equal in number to any one subsampling level, but is often equal to a subsampling level of 0.1 or 0.05. We compare down-sampled lexicase selection with subsampling level of 0.1 to lexicase selection using a static set of 10 cases in Table~\ref{table:static}. Down-sampled lexicase performed significantly better on 11 of the 12 problems tested. This gives strong evidence for the importance of randomly changing the subsample each generation, which was the conclusion also found by~\cite{Ferguson:2019:GPTP}.

\section{Hypotheses for Down-sampled Lexicase Selection's Performance}

All of our results point to the considerable benefits of down-sampled lexicase selection compared to standard lexicase selection. Additional evidence comes from a recent benchmarking of parent selection techniques for program synthesis, which found down-sampled lexicase selection to perform best out of a field of 21 parent selection techniques~\citep{Helmuth:2020:GECCO:benchmarking}. We therefore turn to the question of what makes down-sampled lexicase selection better than other parent selection methods. In this section, we present three distinct hypotheses examining the origins of the benefits bestowed by down-sampled lexicase selection, and conduct experiments to provide evidence for or against these hypotheses.

\subsection{Hypothesis: Depth of Search}

It seems clear that a primary (and possibly the only) benefit of down-sampled lexicase selection is that it allows GP to consider more individuals (i.e. points in the search space) within the same budget of program executions. \cite{Ferguson:2019:GPTP} argue in particular that ``deeper evolutionary searches'', i.e. having a larger maximum number of generations, leading to longer lineages of evolution, is responsible for improvements in performance---we call this the \textit{generations hypothesis}. We present a competing hypothesis, the \textit{search space hypothesis}, that down-sampled lexicase selection's better performance is simply due to evaluating a larger number of individuals, but not related to the depth of the search.

To test these hypotheses, we devised an experiment in which we use down-sampled lexicase selection, but instead of increasing the maximum number of generations per run, we increase the population size while maintaining a fixed number of program executions. For example, with a subsampling level of 0.25, we will increase the population size by 4 times, from 1000 to 4000. This experiment will have GP evaluate the same number of points in the search space as using an increased maximum generations, but will not allow for longer evolutionary lineages than standard lexicase selection, as each run is limited to 300 generations. We test three representative subsampling levels for increased population size, and compare them to the equivalent subsampling levels with increased maximum generations, using the same data as Table~\ref{table:down-sample}.


\begin{table}[t]
    \centering
    \caption{Number of successes out of 100 GP runs of down-sampled lexicase selection at three different subsampling levels. This compares increasing population size to increasing maximum generations, with the latter being identical to the data in Table~\ref{table:down-sample}. Underlined results are significantly better than the corresponding results at the same subsampling level using a chi-squared test. The ``Mean Rank'' gives the average rank of each of the six treatments, so that ranks vary from 1 to 6.}
    \label{table:down-sample-increase-population}
    \rowcolors{2}{gray!15}{white}
	\begin{tabular} {l rrr | rrr}
    	\toprule
    	 & \multicolumn{3}{c|}{Population} & \multicolumn{3}{c}{Generations} \\
        Problem                    & 0.05 & 0.1 & 0.25 & 0.05 & 0.1 & 0.25 \\
        \midrule
        Compare String Lengths     & 48   & 32      & 42 & 38 & 25 & 51    \\
        Double Letters             & 53   & 42   & 35 & \underline{87} & \underline{72} & \underline{50}    \\
        Last Index of Zero         & 76   & 72    & 77 & 72 & 68 & 65    \\
        Mirror Image               & 100  & 100   & 100 & 100 & 99 & 99    \\
        Negative To Zero           & 86   & 86    & 91 & 84 & 86 & 82    \\
        Replace Space With Newline  & 99   & 100    & 95 & 99 & 96 & 100     \\
        Scrabble Score             & 18   & \underline{50}  & \underline{64} & 18 & 19 & 31     \\
        Smallest                   & 99   & 100   & 100 & 100 & 99 & 98    \\
        String Lengths Backwards   & 100  & 100     & 98 & 99 & 96 & 95     \\
        Syllables                  & 24   & 55     & 76 & \underline{48} & 61 & 64    \\
        Vector Average             & 100  & 93    & 99 & 100 & 98 & 97    \\
        X-Word Lines               & 94   & 96    & 84 & 98 & 95 & 91    \\
        \midrule
        Mean    & 74.7 & 77.2 & 80.1  & 78.6 & 76.2 & 76.9  \\
        Mean Rank   &   3.2	&  3.4  &	3.2	&   3.3	&  4.0 	&   4.0 \\
        \bottomrule
    \end{tabular}
\end{table}

We present results using down-sampled lexicase selection with increased population sizes in Table~\ref{table:down-sample-increase-population}. We compare results at the same subsampling level between increased generations and increased population sizes. Out of the 36 comparisons, 2 sets of runs were significantly better with increased population, and 4 were significantly worse. The mean 
success rates across problems are comparable to those with increased generations. We additionally present the average ranking of 6 down-sampled lexicase selection methods (3 that increase population size and 3 that increase maximum generations) across 10 of the problems, excluding the easy problems Mirror Image and Smallest, for which differences only reflect minor differences in generalization rate. The average ranks are all quite close to the overall average rank of 3.5, with increased population having a slightly better average rank across the three subsampling levels, 3.3 versus 3.8.

We take these results as evidence against the generations hypothesis, in that increasing population size while fixing the maximum number of generations produces very similar performance to increasing generations. These results give credence to the search space hypothesis, that we only need to have down-sampled lexicase selection increase the number of individuals we evaluate during evolution, whether that increase comes from increases in population size or more generations. While these conclusions reflect the general results, there are some interesting problem-specific trends to note in Table~\ref{table:down-sample-increase-population}. Increasing generations produced significantly better results on the Last Index of Zero problem at all three subsampling levels, and the inverse was true on Scrabble Score for two of the three subsampling levels. Keeping this in mind, we recommend utilizing the bonus program evaluations allowed by down-sampling on increasing the maximum generations or population size, as both lead to similarly good performance; the choice between the two may come down to other factors within the GP system or to a particular problem.

\subsection{Hypothesis: Changing Environment}

One interesting aspect of down-sampled lexicase selection is that it changes the set of subsampled training cases every generation. If we think of the set of training cases as the challenges encountered by each individual, this corresponds to an environment that changes over time, requiring the evolving population to adapt to new circumstances (i.e. cases). In contrast, with a fixed set of training cases, lexicase selection provides a static environment, though one in which individuals encounter challenges in a different order for each selection. Changing environments often have interesting effects on evolutionary dynamics~\citep{levins1968evolution}, and empirical studies of evolving populations of {\it Saccharomyces cerevisiae} yeast~\citep{bhs21}, logic functions~\citep{Kashtan13711}, and digital organisms~\citep{Nahum,cwo19} have  demonstrated that the speed and effectiveness of adaptive evolution can be affected, and in some cases enhanced, by environmental variation. This led us to ask whether environmental variation might be responsible for the benefits of down-sampled lexicase selection. Here we explore the hypothesis that down-sampled lexicase selection changes the evolutionary dynamics in a positive way beyond increasing the number of individuals that are evaluated.

To test this hypothesis, we designed an experiment that uses a static set of training cases like with lexicase selection, but has each selection only use a subsample of those cases, like with down-sampled lexicase selection. In particular, we use \textit{truncated lexicase selection}, which evaluates every individual on every training case each generation, but cuts off each lexicase selection after using a fixed number of cases~\citep{Spector:2017:GPTP}. In our experiment, we compare down-sampled lexicase selection at the 0.1 subsampling level with truncated lexicase selection also using only 10\% of the cases for each selection. The main difference between the two is that across all selections, truncated lexicase selection uses every training case each generation, where down-sampled lexicase selection uses the same subsample for every selection.\footnote{\cite{Ferguson:2019:GPTP} also conduct an experiment comparing truncated lexicase selection to down-sampled lexicase selection, but to address a different question; we see no contradiction between their results and the ones we present here.}

In our experiment, we run both down-sampled lexicase and truncated lexicase selections for 3000 generations. As truncated lexicase selection requires every individual to be evaluated on every training case each generation, this is not a fair comparison in terms of total program executions, but it is not meant to be. If the ``changing environments'' hypothesis holds, then down-sampled lexicase selection should produce better results than truncated lexicase selection, since its environment changes each generation where truncated lexicase selection's does not. We chose three problems for which down-sampled lexicase selection performed much better than standard lexicase selection over 300 generations, ensuring there is a possibility of performing worse than down-sampled lexicase selection.

\begin{table}[t]
    \centering
    \caption{Number of successes out of 100 GP runs of down-sampled lexicase and truncated lexicase selections, both at the 0.1 level, and both over 3000 generations. Underlined results are significantly better using a chi-squared test.}
    \label{table:truncated}
    \rowcolors{2}{gray!15}{white}
	\begin{tabular} {l rr}
    	\toprule
        Problem        & Down-sampled & Truncated \\
        \midrule
        Double Letters & 72          & 69        \\
        Scrabble Score & 19          & \underline{90}        \\
        Vector Average & 98          & 100       \\
        \bottomrule
    \end{tabular}
\end{table}

Table \ref{table:truncated} presents the number of successful runs of down-sampled lexicase selection and truncated lexicase selection with a maximum of 3000 generations. Over these three problems, truncated lexicase selection performed significantly better than down-sampled lexicase selection on the Scrabble Score problem, and very similarly on the other two problems. So, not only was down-sampled lexicase selection not better, it was a bit worse. This gives some evidence against the hypothesis that the ``changing environment'' of down-sampled lexicase selection contributes to its success, though we admit that there may be other beneficial evolutionary dynamics at play not captured by this experiment. We also want to emphasize that this experiment does not suggest that truncated lexicase selection should be preferred over down-sampled lexicase selection, or even standard lexicase selection for that matter; truncated lexicase selection used 10 times as many program executions in these runs as down-sampled lexicase selection, meaning they are not being compared on a level playing field.

\subsection{Hypothesis: Better Generalization}

\begin{table}[tbp]
    \centering
    \caption{Comparing generalization rates of lexicase selection and down-sampled lexicase selection with a subsampling level of 0.25. These generalization rates are for the success rates in Table~\ref{table:lex-down-sample}. None of the differences in generalization were significant.}
    \label{table:lex-down-sample-generalization}
    \rowcolors{2}{gray!15}{white}
    \begin{tabular}{l rr}
	\toprule
    Problem                & Down-sampled      & Lexicase \\
    \midrule
    Checksum                   & 1.00                & 1.00                \\
    CSL                                & 0.61       & 0.49                 \\
    Count Odds                 & 1.00                & 1.00                \\
    Digits                               & 0.60      & 0.66                \\
    Double Letters                        & 0.98      & 0.95               \\
    Even Squares                           & 1.00       & -                \\
    For Loop Index                     & 1.00       & 0.67                 \\
    Grade                                & 1.00      & -                   \\
    Last Index of Zero         & 0.66                & 0.67                \\
    Median                                & 0.69      & 0.57               \\
    Mirror Image                        & 0.99    & 1.00                   \\
    Negative To Zero           & 0.83                & 0.84                \\
    Number IO                  & 0.99                & 0.98                \\
    Pig Latin                  & -                   & -                   \\
    RSWN                       & 1.00                & 1.00                \\
    Scrabble Score                      & 1.00      & 0.93                 \\
    Small Or Large                          & 0.42    & 0.32               \\
    Smallest                             & 0.98       & 1.00               \\
    String Differences                     & 1.00     & -                  \\
    SLB                        & 1.00                & 1.00                \\
    Sum of Squares             & 1.00                & 1.00                \\
    Super Anagrams                        & 1.00     & 0.80                \\
    Syllables                             & 0.96     & 0.97                \\
    Vector Average             & 1.00                & 1.00                \\
    Vectors Summed                        & 0.95    & 0.92                 \\
    X-Word Lines                         & 0.98    & 1.00                   \\
    \bottomrule
    \end{tabular}
\end{table}

As discussed in the Related Work section above, down-sampling has been used (without lexicase selection) in both GP and machine learning more broadly as a method to combat overfitting and increase the generalization of solutions. There is plenty of room for improvement in generalization on some of our benchmark problems, with 6 problems having generalization rates below 0.7 when using lexicase selection. Does down-sampling improve generalization when using lexicase selection?

All of our successful run counts above only include generalizing solutions that pass a large set of random, unseen test cases. We look at the proportion of solution programs that pass the training set that also pass the test set to calculate the \textit{generalization rate} for each set of runs. For the extended set of 26 benchmark problems presented in Table~\ref{table:lex-down-sample}, we present the generalization rate for each problem in Table~\ref{table:lex-down-sample-generalization}. Even though there are some minor differences in generalization between lexicase and down-sampled lexicase selections, none of them are significantly different using a chi-squared test. Problems that appear to have a large gap between the two, such as For Loop Index and Super Anagrams, do not have enough solutions to show significance.

At this point we have no evidence to suggest that down-sampling improves lexicase selection's generalization rate. In fact, down-sampled lexicase selection displays poor generalization on many of the same problems that lexicase selection does. Thus we cannot attribute the improved performance of down-sampled lexicase selection to avoiding overfitting and improving generalization.



\section{Conclusions}

In this paper we have shed more light on the performance and mechanisms of down-sampled lexicase selection. We conducted more extensive benchmarking of down-sampled lexicase selection than has been conducted before, finding that it performs well across a large range of benchmark problems and subsampling levels.
We describe some of the drawbacks of using very low subsampling levels, despite their ability to produce competitive problem solving performance.
We find that it is important to change training cases every generation within a larger set of training cases, as a subsampling method that uses a static set of cases throughout evolution performed much worse than down-sampled lexicase selection.

We then considered the hypothesis that down-sampled lexicase selection performs well because of its ability to search for more generations, leading to deeper evolutionary lineages. Our experiment that makes use of down-sampled lexicase selection's extra program executions to increase the population size rather than extending evolutionary time provides evidence against this hypothesis, since approximately the same benefit is obtained with larger populations as with more generations. We also examine the hypothesis that down-sampled lexicase selection's changing of training cases every generation acts like an environment changing over evolutionary time, contributing to its success. Our experiment using truncated lexicase selection provides evidence against this hypothesis, though other environmental effects could be at play. A third experiment showed that down-sampled lexicase selection does not produce better generalization rates of solution programs compared to lexicase selection, despite this being a benefit of down-sampling in other machine learning systems. These experiments lead us to believe that the primary cause of down-sampled lexicase selection's success is that it allows evolution to consider more programs throughout evolution.

This work and that of \cite{Ferguson:2019:GPTP} and \cite{Hernandez:2019:GECCOcomp} use problems from the same general program synthesis benchmark suite. We would certainly like to see similar experiments performed in other problem domains, where training set subsampling has been used previously, but not to our knowledge in conjunction with lexicase selection.

This research points to the importance of maximizing the number of points in the search space---individuals---that genetic programming considers throughout evolution. In this paper we push the abilities of down-sampled lexicase selection to increase the number of individuals considered to the extreme, and find that at the 0.01 and 0.02 subsampling levels, problem-solving performance remains surprisingly good, while actual processor performance diminishes. We would be interested to see what effects such low subsampling levels have on population dynamics such as diversity, considering they allow lexicase to only select a tiny fraction of the individuals in the population.

Other methods that increase the number of individuals considered by genetic programming without sacrificing information about individuals' performances (or even ones that do sacrifice some information, as in down-sampled lexicase selection) could provide additional benefits. Exploring this avenue illuminated by down-sampled lexicase selection may yield other techniques that, possibly in combination with down-sampled lexicase selection, could continue to drive the field forward.

\section{Acknowledgements}

We thank Emily Dolson, Amr Abdelhady, and the Hampshire College Computational Intelligence Lab for discussions that improved this work.
This material is based upon work supported by the National Science Foundation under Grant No. 1617087. Any opinions, findings, and conclusions or recommendations expressed in this publication are those of the authors and do not necessarily reflect the views of the National Science Foundation.

\footnotesize
\bibliographystyle{apa-good}
\bibliography{down-sampled} 

\end{document}